%% file: acl_latex.tex
\pdfoutput=1
 
\documentclass[11pt]{article}
 
\usepackage[final]{acl}
 
\usepackage{times}
\usepackage{latexsym}
 
\usepackage{pifont}
 
\usepackage[T1]{fontenc}
 
\usepackage[utf8]{inputenc}
 
\usepackage{microtype}
 
\usepackage{inconsolata}
 
\usepackage{graphicx}
 
\usepackage{microtype}
\usepackage{booktabs} 
\usepackage{hyperref}
\usepackage{tabularx}

\usepackage{amsmath}
\usepackage{amssymb}
\usepackage{mathtools}
\usepackage{amsthm}
\usepackage{caption}
\usepackage{subcaption}
\usepackage{wrapfig}
\usepackage{bm}
\usepackage{bbm}
\usepackage{url}
\usepackage{multirow}
\usepackage{enumitem}
\usepackage{diagbox}
\PassOptionsToPackage{table}{xcolor}
\usepackage{color,colortbl}
\usepackage{xcolor}
\usepackage{makecell}
\usepackage[most]{tcolorbox}
\usepackage{booktabs}
\usepackage{array}
\usepackage{tabularx}
\newcolumntype{Y}{>{\raggedright\arraybackslash}X}
\newcolumntype{C}[1]{>{\centering\arraybackslash}p{#1}}
\definecolor{COLOR_MEAN}{HTML}{f0f0f0}
 
\newtcolorbox[list inside=prompt]{prompt}[1][]{
    enhanced,
    colback=green!5!gray!10,
    colframe=green!20!gray!55,
    colbacktitle=green!20!gray!55,
    coltitle=white,
    fonttitle=\bfseries\sffamily,
    fontupper=\small\ttfamily,
    title=Prompt,
    attach boxed title to top center={yshift=-3pt},
    boxed title style={
        rounded corners=northeast,
        rounded corners=southeast,
        boxrule=0pt
    },
    rounded corners,
    boxrule=1.5pt,
    left=2pt,          %
    right=0pt,         %
    top=5pt,           %
    bottom=0pt,        %
    boxsep=2pt,        %
    before skip=5pt,   %
    after skip=5pt,    %
    breakable,
    #1,
}
 
\definecolor{greyblue}{RGB}{120,145,180}
\newtcolorbox[list inside=prompt]{response}[1][]{
    enhanced,
    colback=greyblue!15,
    colframe=greyblue!60,
    colbacktitle=greyblue!60,
    coltitle=white,
    fonttitle=\bfseries\sffamily,
    fontupper=\small\ttfamily,
    title=Prompt,
    attach boxed title to top center={yshift=-3pt},
    boxed title style={
        rounded corners=northeast,
        rounded corners=southeast,
        boxrule=0pt
    },
    rounded corners,
    boxrule=1.5pt,
    left=2pt,          %
    right=0pt,         %
    top=5pt,           %
    bottom=0pt,        %
    boxsep=2pt,        %
    before skip=5pt,   %
    after skip=5pt,    %
    breakable,
    #1,
}
 
\usepackage{pgfplots}
\pgfplotsset{compat=1.18}
\usepgfplotslibrary{groupplots}
 
\definecolor{baseColor}{HTML}{888780}
\definecolor{oursColor}{HTML}{534AB7}
\definecolor{deltaColor}{HTML}{0F6E56}
\definecolor{rowShade}{HTML}{EEECFB}

\title{TRN-R1-Zero: Text-rich Network Reasoning via LLMs with Reinforcement Learning Only}

\author{Yilun Liu, Ruihong Qiu \and Zi Huang \\
        School of Electrical Engineering and Computer Science \\
        The University of Queensland \\
        Brisbane, Queensland, Australia \\
        \texttt{\{yilun.liu, r.qiu, helen.huang\}@uq.edu.au}}

\begin{document}
\maketitle
\begin{abstract}
Zero-shot reasoning on text-rich networks (TRNs) remains a challenging frontier, as models must integrate textual semantics with relational structure without task-specific supervision. While graph neural networks rely on fixed label spaces and supervised objectives, recent large language model (LLM)-based approaches often overlook graph context or depend on distillation from larger models, limiting generalisation. We propose TRN-R1-Zero, a post-training framework for TRN reasoning trained solely via reinforcement learning. TRN-R1-Zero directly optimises base LLMs using a Neighbour-aware Group Relative Policy Optimisation objective that dynamically adjusts rewards based on a novel margin gain metric for the informativeness of neighbouring signals, effectively guiding the model toward relational reasoning. Unlike prior methods, TRN-R1-Zero requires no supervised fine-tuning or chain-of-thought data generated from large reasoning models. Extensive experiments across citation, hyperlink, social and co-purchase TRN benchmarks demonstrate the superiority and robustness of TRN-R1-Zero. Moreover, relying strictly on node-level training, TRN-R1-Zero achieves zero-shot inference on edge- and graph-level tasks, extending beyond cross-domain transfer. The codebase is publicly available at \href{https://github.com/superallen13/TRN-R1-Zero}{https://github.com/superallen13/TRN-R1-Zero}.
\end{abstract}

\section{Introduction}
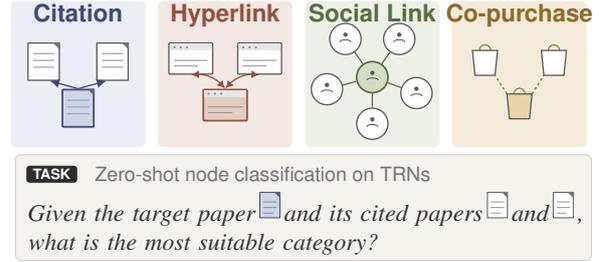
\begin{figure}[!t]
    \centering
    \input{figures/trn.tex}
    \caption{\textbf{Top}: Examples of text-rich networks (TRNs) from citation, hyperlink, social and co-purchase domains.
    \textbf{Bottom}: An example of reasoning-based user query over TRNs.}
    \label{fig:tag_domains}
\end{figure}

Text classification is a cornerstone task in natural language processing, underpinning applications from information retrieval to content recommendation. Yet, in real-world scenarios, texts seldom exist in isolation: scientific papers cite one another, Wikipedia pages are interlinked through hyperlinks, users in social networks follow each other, and e-commerce products often co-occur in purchases. These relational connections naturally form text-rich networks (TRNs), where nodes correspond to textual entities and edges capture their semantic or functional relations. As illustrated in Figure~\ref{fig:tag_domains}, TRNs from citation, hyperlink, social, and co-purchase domains exhibit rich relational structures that go beyond isolated document understanding~\cite{caselink, casegnn, lexa, LocAgent, graph-plan, tntood, cat, PUMA, gcondenser}. Effectively reasoning over such TRNs, particularly in zero-shot classification settings without domain-specific supervision, is a crucial step toward more generalisable and context-aware language intelligence.

Existing large language model (LLM)-based methods for node classification on TRNs generally follow two paradigms. (1) Encoder-based approaches use LLMs as text encoders for both node and label descriptions~\cite{zerog, uniglm, LLM-BP}. The resulting embeddings are aggregated through structure-aware mechanisms across neighbouring nodes, and classification is performed via node-label similarity. However, these methods largely treat the LLM as a feature extractor rather than an explicit reasoner. (2) Generative approaches, on the other hand, reformulate node classification as a label-token generation task. To incorporate structural information, some employ soft-embedding techniques that align graph encodings with the LLM’s embedding space~\cite{graphgpt, tea-glm, llaga, gofa}, while others use natural language descriptions of graph structures as inputs~\cite{nlgraph, graphwiz, graphadapter, glbench, nodebed}. Recent work on both text-based or non-text networks attempt to transfer reasoning abilities from large reasoning models (LRMs) by fine-tuning on chain-of-thought data~\cite{graphwiz, graph-theory-r1, graph-r1}. Despite these advances, \textbf{existing paradigms struggle to directly elicit explicit reasoning within LLMs on TRNs}, while often relying on additional supervision or external reasoning resources.

\begin{table}[!t]
    \centering
    \resizebox{\linewidth}{!}{
    \begin{tabular}{lccc}
        \toprule
        Model & LRM & CoT SFT & Reason. \\
        \midrule
        GraphWiz & \ding{51} (GPT-4) & \ding{51} (GPT-4) & \ding{51} \\
        Graph-NPH-R1 & \ding{55} & \ding{51} (QwQ-32B) & \ding{51} \\
        Graph-R1   & \ding{51} (DeepSeek-V3) & \ding{51} (DeepSeek-R1) & \ding{51} \\
        \midrule
        TRN-R1-Zero & \ding{55} & \ding{55} & \ding{51} \\
        \bottomrule
    \end{tabular}
    }
    \caption{Comparison of reasoning training for LLMs on graph tasks. For graph theory problems, GraphWiz~\cite{graphwiz} and Graph-NPH-R1~\cite{graph-theory-r1} rely on mimicking the CoT process of larger LLMs. For text-based network problems, Graph-R1~\cite{graph-r1} and GraphWiz further depend on external LRMs to provide reasoning supervision. Our TRN-R1-Zero achieves reasoning ability without relying on LRMs or their generated CoT data.}
    \label{tab:method_comp}
\end{table}

To address these limitations, we propose TRN-R1-Zero, a reinforcement learning-based framework that enables explicit reasoning on TRNs without any supervised fine-tuning or external distillation. Unlike encoder-based methods that treat LLMs merely as feature extractors, or generative approaches that depend on pre-generated reasoning traces, TRN-R1-Zero learns to reason relationally through direct optimisation over the underlying graph context. We develop a novel Neighbour-aware Group
Relative Policy Optimisation objective with a margin gain metric, which leverages local neighbourhood information as adaptive signals to guide reasoning training, allowing it to effectively infer structural and semantic relationships for node classifications on text-rich networks in unseen domains. This design activates the LLM’s reasoning capability intrinsically, rather than relying on external supervision or task-specific data. A comparative summary of existing paradigms and TRN-R1-Zero is provided in Table~\ref{tab:method_comp}. Our main contributions are:
\begin{enumerate}
\item An RL-only pipeline for zero-shot node classification on TRNs, without distillation, SFT, or external LRMs.
\item A neighbour-aware policy objective with a margin gain mechanism that explicitly encourages the use of relational context.
\item Extensive experiments on citation, hyperlink, social, and co-purchase TRNs demonstrate consistent zero-shot gains in cross-domain and cross-task settings over prior methods.
\end{enumerate}

\section{Related Work}

\paragraph{Large Language Models for Node Classification.}
Existing approaches to zero-shot node classification on text-rich networks (TRNs) can be categorised into encoder-based and generative paradigms.  

\textbf{Encoder-based} methods use language models (LMs) or LLMs primarily as text encoders, generating embeddings for nodes and labels that are subsequently aligned or aggregated by external algorithms. 
ZeroG~\cite{zerog} fine-tunes Sentence-BERT~\cite{sbert} with LoRA~\cite{lora} to effectively encode both node texts and label descriptions. 
UniGLM~\cite{uniglm} fine-tunes BERT~\cite{bert} into a more generalised text encoder through contrastive learning, boosting the downstream graph model performance trained in this embedding space. 
TAPE~\cite{tape} fine-tunes DeBERTa~\cite{deberta} with explanations and predictions generated from an extra LLM.
Nevertheless, these fine-tuned encoders exhibit limited generalisation ability because of the small model size and data scarcity of the tuning phase.
LLM-BP~\cite{LLM-BP} employs LLM2Vec~\cite{llm2vec} as a text encoder and applies propagation-based techniques to integrate neighbour information. These methods, however, fail to exploit the explicit reasoning capabilities of LLMs.  

\textbf{Generative} methods formulate node classification as a text generation task. GraphGPT~\cite{graphgpt}, GOFA~\cite{gofa}, TEA-GLM~\cite{tea-glm}, and LLaGA~\cite{llaga} employ a learnable mapping model to project graph structures into the LLM’s token embedding space, creating soft embeddings that enable the LLM to generate graph-aware representations after supervised fine-tuning.
Alternatively, other works describe graph information directly in natural language, enabling the LLM to understand over structural context without explicit graph encoders~\cite{graphadapter, glbench, nodebed}.

\paragraph{Large Language Model Reasoning.}
LLMs trained with reinforcement learning have demonstrated impressive reasoning abilities and human-like performance across a range of tasks, including mathematics~\cite{grpo, math-arena}, task planning~\cite{cos}, code generation~\cite{swe-ben}, and debugging~\cite{ldb}.
Proximal Policy Optimisation (PPO)\cite{ppo} serves as the foundation for reasoning-oriented fine-tuning.
The recent Group Relative Policy Optimisation (GRPO)\cite{grpo, deepseek-r1} introduces a rule-based objective that enables reasoning skills without human-annotated supervision, while Dr.GRPO~\cite{drgrpo} further enhances reward shaping and variance control during reasoning training.

The \textbf{reasoning ability of LLMs has recently been extended to structured data}.
For general graphs without textual attributes, GraphWiz~\cite{graphwiz} and Graph-NPH-R1~\cite{graph-theory-r1} leverage large reasoning models (LRMs) to generate chain-of-thought (CoT)\cite{cot} data for reasoning over graph-theoretic problems such as shortest path and connectivity.
Graph-R1\cite{graph-r1} further targets text-rich graphs, using LRMs to produce long CoT traces that supervise the fine-tuning of smaller models.
In contrast, TRN-R1-Zero removes the need for distillation or externally generated CoT data from larger LRMs, directly eliciting reasoning ability within the base model itself.

\section{Methodology: TRN-R1-Zero}
\begin{figure*}
    \centering
    \includegraphics[width=1.0\linewidth]{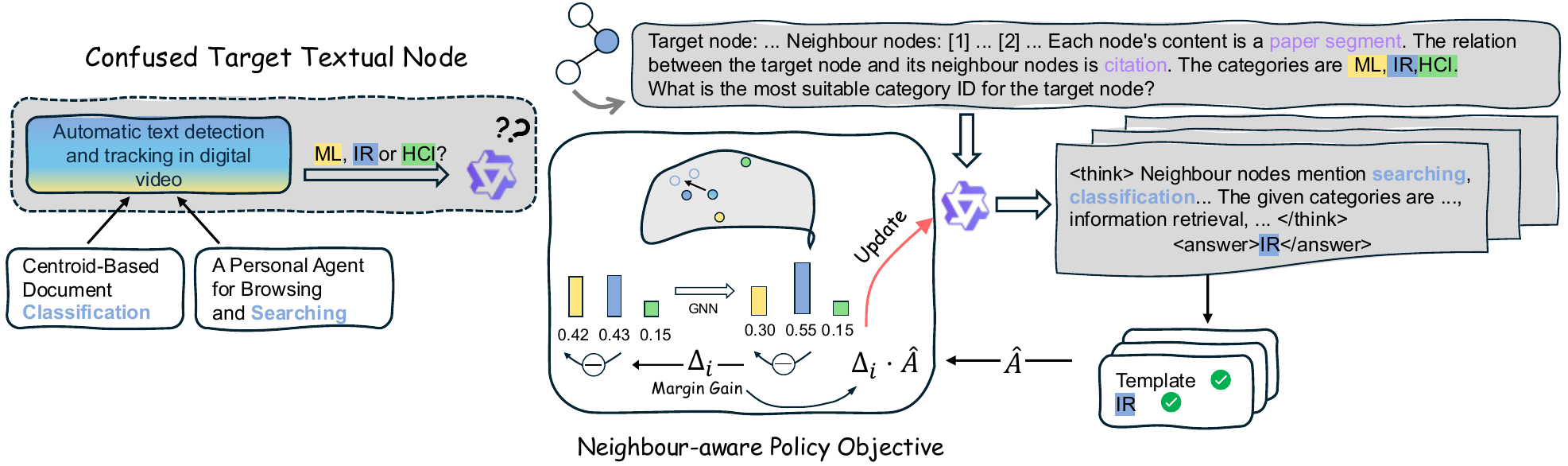}
    \caption{Overall training pipeline of TRN-R1-Zero, comprising three key components:
    graph sampling, prompt construction, and neighbour-aware policy objective.}
    \label{fig:framework}
\end{figure*}

To perform node classification with reasoning on text-rich networks, this section describes how TRN-R1-Zero achieves this capability through a novel optimisation paradigm, as illustrated in Figure~\ref{fig:framework}.

\subsection{Zero-shot Node Classification}

Given a text-rich network $G = (V, E, Y)$, where $V = \{ v_1, \dots, v_{|V|} \}$ denotes the set of nodes with associated texts, $E \subseteq V \times V$ denotes the set of edges, and $Y = \{ y_1, \dots, y_{|Y|} \}$ denotes the set of label texts, the objective is to predict the label of a target node $v_i \in V$ without any supervision from the target network.

\paragraph{Classification as Token Generation.}
Given a large language model $\mathcal{M}_\theta$, the input comprises the text of the target node $t_i$, its neighbourhood $\mathcal{N}(v_i)$, and the candidate label texts $Y$. Each class $y \in Y$ is mapped to a discrete identifier token (e.g., ``1'', ``2'', ``3''). Thus, node classification is reformulated as a next-token prediction task:
\begin{equation}
\hat{y}_i = \arg\max_{y \in Y} P_\theta\big( y \mid \mathcal{P}(t_i, \mathcal{N}(v_i), Y) \big),
\end{equation}
where $\mathcal{P}(\cdot)$ denotes the constructed prompt that integrates node, neighbour, and label information.

\subsection{Prompt with Neighbourhood Sampling}

The input to the LLM is constructed by combining the target node text, sampled neighbour texts, and candidate label descriptions into an instruction-style prompt (see Box~1 below).  
In our prompt design, neighbourhood sampling serves a dual purpose: it controls the input length and acts as a form of data augmentation.  
For each target node $v_i$, multiple subgraphs are randomly sampled following a fixed width--depth strategy, where (i) \textbf{width} limits the number of included neighbours, and (ii) \textbf{depth} truncates the text of each neighbour.  
By repeatedly drawing diverse subsets of neighbours, the LLM is exposed to varied local contexts, effectively expanding the training corpus and mitigating overfitting in low-resource graph settings.

\begin{center}
    \begin{prompt}[title={Box 1: Train Prompt for TRN-R1-Zero}, label=tab:prompt_template]
    \# System Prompt
    
    You are a helpful AI Assistant that provides well-reasoned and detailed responses. 
You first think about the reasoning process as an internal monologue and then provide the user with the answer. 
Respond in the following format: \\
\texttt{<think>} \\
\texttt{...} \\
\texttt{</think>} \\
\texttt{<answer>} \\
\texttt{...} \\
\texttt{</answer>} \\

\# Graph Prompt

Target node: \texttt{\{target\_node\_text\}} \\
Neighbour nodes: \texttt{\{neighbor\_node\_text\}} \\

\# Task Instruction

I provide the content of the target node and its neighbour nodes. Each node content is 
\texttt{\{node\_type\}}. The relation between the target node and its neighbour nodes is 
\texttt{\{relation\}}. The \texttt{\{num\_categories\}} categories are: \texttt{\{labels\}}. \\
Question: Based on the information of the target and neighbour nodes, 
predict the category ID (0 to \texttt{\{max\_id\}}) for the target node.
\end{prompt}
\end{center}

\subsection{Neighbour-aware Group Relative Policy Optimisation Objective}

Reinforcement learning for LLM post-training builds upon the GRPO objective~\cite{grpo}, 
a variant of PPO~\cite{ppo} adapted for sequence-level rewards.  
Given a query $q$ and an output sequence $o = (o_1, \dots, o_{|o|})$,  
the objective is defined as:
\begin{align}
\label{eq:grpo}
\textstyle
\mathcal{J}(\theta) =
\mathbb{E}_{q \sim \mathcal{D},\; o \sim \pi_{\theta_{\text{old}}}} \Bigg[
\sum_{t=1}^{|o|}
\min\!\Big(
r_t \hat{A}_t, \notag\\
\text{clip}(r_t, 1-\epsilon, 1+\epsilon)\,\hat{A}_t
\Big)\Bigg].
\end{align}
where 
$r_t = \tfrac{\pi_\theta(o_t \mid q,o_{<t})}{\pi_{\theta_{\text{old}}}(o_t \mid q,o_{<t})}$ 
is the token-level importance sampling ratio, 
and $\hat{A}_t$ is the advantage estimator. 
\begin{equation}
    \hat{A}_t = \tfrac{R_i - \bar R}{\mathrm{std}(R)}.
\end{equation}
A KL regularisation term
\begin{equation}
-\beta \cdot \mathrm{KL}\!\left[\pi_\theta(\cdot \mid q,o_{<t}) \,\|\, \pi_{\text{ref}}(\cdot \mid q,o_{<t})\right]
\end{equation}
is added to penalise deviation from a frozen reference policy $\pi_{\text{ref}}$, stabilising optimisation.

This formulation can hinder reward shaping, as the standard deviation term dampens variations in $R_i$. 
Dr.GRPO~\cite{drgrpo} addresses this by removing the denominator:
\begin{equation}
\hat{A}_t = R_i - \bar R.
\end{equation}
This allows shaped rewards to influence optimisation magnitude directly. 
Although Dr.GRPO is commonly implemented without KL, empirical results in this task reveal that omitting KL causes unstable training. 
Therefore, the adopted objective is Dr.GRPO with KL regularisation, which preserves both stability and effective scaling.

\paragraph{Margin Gain: Quantifying Neighbouring Contribution.}\label{sec:margin-gain}
During reasoning over neighbouring nodes, the neighbourhood information may either complement or distract from the target node’s textual signal.
To identify cases where neighbour information plays a pivotal role,
we introduce a margin gain metric that quantifies how much the classification decision boundary shifts after incorporating neighbours.

Let $e_i = f(x_i) \in \mathbb{R}^d$ be the embedding of node text $x_i$ and $e_c = f(y_c)$ the embedding of label text $y_c$, 
where $f(\cdot)$ is a frozen text encoder.  
The raw logit of node $i$ for class $c$ is
\begin{equation}
\ell_{i,c} = e_i^\top e_c .
\end{equation}
Let $y_i$ denote the ground-truth class of node $i$. 
The raw margin score is defined as
\begin{equation}
m_i(\ell) = \ell_{i,y_i} - \max_{c \neq y_i} \ell_{i,c},
\end{equation}
which measures how confidently the encoder classifies the node text in isolation.

To measure the influence of neighbours, we apply a lightweight $K$-layer Simple Graph Convolution (SGC)-style aggregator~\cite{sgc}:
\begin{equation}
\label{eq:sgc}
\tilde{E} = \big(D^{-\tfrac{1}{2}} A D^{-\tfrac{1}{2}}\big)^K E,
\end{equation}
where $A$ is the adjacency matrix with self-loops, $D$ is its degree matrix, 
and $E$ stacks all node embeddings row-wise. 
The aggregated embedding $\tilde e_i$ induces aggregated logits
\begin{equation}
\tilde{\ell}_{i,c} = \tilde{e}_i^\top e_c ,
\end{equation}
and a corresponding aggregated margin
\begin{equation}
m_i(\tilde{\ell}) = \tilde{\ell}_{i,y_i} - \max_{c \neq y_i} \tilde{\ell}_{i,c}.
\end{equation}

Therefore, the margin gain can be defined to quantify the contribution of the neighbourhood:
\begin{equation}
\label{eq:margin-gain}
\Delta_i = m_i(\tilde{\ell}) - m_i(\ell),
\end{equation}
which captures how much neighbourhood aggregation improves (or degrades) the classification margin.

Intuitively, $\Delta_i > 0$ indicates that neighbours provide helpful context; 
$\Delta_i \approx 0$ suggests neighbourhood information is redundant; 
and $\Delta_i < 0$ implies neighbours are distracting. 
We use the absolute value $|\Delta_i|$ to measure the strength of neighbourhood influence, 
regardless of whether the effect is positive or negative.

\paragraph{Reward Design with Margin Gain.}\label{sec:reward-shaping}
Reinforcement learning with GRPO assigns each prompt a scalar reward $R_i$, 
which determines the magnitude of policy gradient updates through the advantage estimator.  
Rather than treating all prompts equally, we scale the rewards by the neighbourhood influence so that samples where neighbours have a stronger impact on decisions receive greater emphasis.

For a rollout $o_i$ associated with node $v_i$, the base reward comprises two components:
\begin{equation}
R_i^{\text{base}} = s_{\text{format}}(o_i) + s_{\text{acc}}(o_i),
\end{equation}
where $s_{\text{format}}$ enforces adherence to the output schema (e.g., correct use of \texttt{<think>} and \texttt{<answer>} tags), 
and $s_{\text{acc}}$ verifies whether the final answer matches the ground-truth identifier token.

To reflect the importance of the neighbourhood via the margin gain from Eq.~(\ref{eq:margin-gain}), 
we define a reshaping factor:
\begin{equation}
\label{eq:reshape}
g_i = \exp(\alpha \cdot |\Delta_i|),
\end{equation}
where $\alpha$ is a temperature hyperparameter controlling sensitivity.  

This exponential form has two intuitive effects:  
(i) when $|\Delta_i| = 0$, $g_i = 1$ and the reward remains unchanged;  
(ii) larger $|\Delta_i|$ values exponentially amplify the reward, encouraging the model to focus more on neighbour-influenced samples during policy optimisation.  
The final reward is therefore:
\begin{equation}
R_i = g_i \cdot R_i^{\text{base}} 
= \exp(\alpha |\Delta_i|)\big(s_{\text{format}}(o_i) + s_{\text{acc}}(o_i)\big).
\end{equation}

Incorporating this reward design into LLM update using objective in Eq.~(\ref{eq:grpo}) will emphasise structurally informative neighbourhoods, 
guiding the LLM generation to more effectively leverage relational context for reasoning on text-rich networks.

\subsection{Inference on Edge and Graph Tasks}
Although TRN-R1-Zero is trained only on node-level tasks, extending it to other TRN tasks such as link prediction and graph reasoning is straightforward. The input prompt requires only the sampled graph with neighbour information, and the task instruction. Detailed prompt designs are provided in Appendix~\ref{app:prompts}, with cross-task experiments reported in Section~\ref{sec:exp-cross-task}.

\begin{table*}[!t]
    \centering
    \vspace*{-8pt}
    \resizebox{1\linewidth}{!}{
    \begin{tabular}{ll|cc|cc|cc|cc|cc}
       \toprule
       \rowcolor{COLOR_MEAN} & &  \multicolumn{2}{c|}{\textbf{Cora}} & \multicolumn{2}{c|}{\textbf{WikiCS}} & \multicolumn{2}{c|}{\textbf{Instagram}}  & \multicolumn{2}{c|}{\textbf{Photo}} & \multicolumn{2}{c}{\textbf{Avg.}}  \\ 
       \rowcolor{COLOR_MEAN} \multirow{-2}{*}{\textbf{Type}} &  \multirow{-2}{*}{\textbf{Method}} & Acc & Macro-F1 & Acc & Macro-F1 & Acc & Macro-F1 &  Acc & Macro-F1  & Acc & Macro-F1 \\  
       \midrule
       \multirow{5}{*}{\textbf{LLM}} 
       & GPT-4o & \colorbox{blue!10}{70.30} & \colorbox{blue!25}{\textbf{71.44}} & 69.69 & 64.51 & 42.42 & 39.79 & \colorbox{blue!25}{\textbf{69.93}} & \colorbox{blue!25}{\textbf{68.55}} & 63.09 & 61.07 \\ 
       & Llama-3.1-8B & 64.55 & 64.41 & 59.43 & 54.16 & 36.98 & 28.32 & 45.49 & 50.44 & 51.61 & 49.33 \\ 
       & Qwen2.5-1.5B-it & 47.96 & 49.91 & 61.71 & 56.17 & 36.82 & 28.37 & 50.72 & 51.50 & 49.30 & 46.49 \\
       & Qwen2.5-7B-it & 67.59 & 67.19 & 67.44 & 63.93 & 52.20 & 50.32 & 55.67 & 59.37 & 60.73 & 60.20 \\
       & Qwen2.5-14B-it & 67.22 & 68.26 & 73.03 & \colorbox{blue!25}{\textbf{70.78}} & \colorbox{blue!25}{\textbf{55.60}} & \colorbox{blue!25}{\textbf{52.94}} & 58.51 & 61.45 & \colorbox{blue!10}{63.59} & \colorbox{blue!10}{63.36} \\
       \midrule
       \multirow{2}{*}{\textbf{GFM}} 
       & ZeroG & 62.55 & 57.56  & 62.71 & 57.87 & 50.71 & 50.43 & 46.27 & 51.52 & 55.56 & 54.35 \\ 
       & LLaGA & 18.82 & 8.49 & 8.20 & 8.29 & 47.93 & 47.70 & 39.18 & 4.71 & 28.53 & 17.30 \\ 
       \midrule
       \multirow{1}{*}{\textbf{SFT + RL}} 
       & Graph-R1 (14B) & 68.15 & 67.34 & \colorbox{blue!10}{73.25} & 70.11 & 52.03 & 52.06 & - & - & - & - \\ 
       \midrule
       \multirow{1}{*}{\textbf{RL Only}}
       & \textbf{TRN-R1-Zero (7B)} & \colorbox{blue!25}{\textbf{72.59}} & \colorbox{blue!10}{70.33} & \colorbox{blue!25}{\textbf{73.63}} & \colorbox{blue!10}{70.30} & \colorbox{blue!10}{54.76} & \colorbox{blue!10}{52.54} & \colorbox{blue!10}{65.12} & \colorbox{blue!10}{64.22} & \colorbox{blue!25}{\textbf{66.53}} & \colorbox{blue!25}{\textbf{64.35}} \\
       \bottomrule
    \end{tabular}
    }
    \caption{Performance comparison under the zero-shot setting with Accuracy ($\%$) and Macro-F1 ($\%$) reported with benchmarks following~\cite{nodebed}. The \colorbox{blue!25}{\textbf{best}} and \colorbox{blue!10}{second-best} results are highlighted per column (paired t-test, $p \leq 0.05$, Bonferroni corrected). Graph-R1 is excluded from the Photo dataset since Photo was used in its pre-training, not qualifying for zero-shot evaluation.}
    \label{tab:zeroshot}
\end{table*}

\section{Experiments}
\subsection{Setup}

\paragraph{Datasets.}
Experiments are conducted on nine datasets spanning four relational structures (citation, hyperlink, social, and co-purchase) and three task types (node-, graph-, and edge-level).
For RL training, \textbf{Citeseer} and \textbf{History} are used to capture citation and co-purchase relations.
The remaining \textbf{Cora}, \textbf{Photo}, \textbf{WikiCS}, and \textbf{Instagram} are used for zero-shot in-domain and cross-domain evaluation, while \textbf{Expla-Graph}~\cite{g-retriever}, \textbf{WikiCS-Link}, and \textbf{Instagram-Link} are used for cross-task evaluation.
All datasets (detailed in Table~\ref{tab:dataset_metadata} and~\ref{tab:datasets}) are sourced from NodeBed~\cite{nodebed}.

\begin{itemize}
    \item \textbf{Cora, Citeseer}: Each node represents a scientific publication including the paper title and abstract. Edges denote citation links between papers, forming a citation network.
    
    \item \textbf{WikiCS}: Each node corresponds to a Wikipedia article, and edges represent hyperlinks between articles, forming a web graph.
    
    \item \textbf{Instagram}: Each node represents a user account, and edges correspond to social-follow relations. Node texts are profile descriptions or short post contents, reflecting social interaction context.
    
    \item \textbf{Photo, History}: Each node corresponds to a product on the Amazon platform. Nodes are customer reviews in Photo and product descriptions in History, and edges capture co-purchase relations between products.

    \item \textbf{Expla-Graph}: Each node denotes a commonsense concept, and each edge denotes the relation between two concepts.

    \item \textbf{WikiCS-Link and Instagram-Link}: Both datasets are constructed from original node-level datasets by retaining the original edges as positives and uniformly sampling an equal number of non-existent edges as negatives.
    
\end{itemize}

\paragraph{Baselines.}
The comparison includes three categories of baselines:
\begin{itemize}
    \item \textbf{LLMs}: GPT-4o is included to represent a potential upper bound of LLM performance.  
    LLaMA-3.1-8B, Qwen2.5-1.5B-Instruct, Qwen2.5-7B-Instruct, and Qwen2.5-14B-Instruct are selected to cover diverse open-source model families and scales.

    \item \textbf{Graph Foundation Models (GFMs)}: ZeroG~\cite{zerog} and LLaGA~\cite{llaga} are evaluated in an intra-domain manner, where each model is pre-trained on the same domain dataset (e.g., arXiv for academic data) before being tested on the target dataset.

    \item \textbf{Reasoning LLMs}: Graph-R1~\cite{graph-r1} introduces a \emph{rethink} template that encourages LLMs to reason carefully and revise their responses before producing the final answer.  
    In the original setup, DeepSeek-v3 is used to summarise node texts into compact representations.  
    For fairness, the following experiments use raw node texts directly.
\end{itemize}

\paragraph{Implementations.}
The dataset statistics are summarised in Table~\ref{tab:datasets}, covering four relation types: citation, co-purchase, hyperlink, and social.  
During training, \textbf{Citeseer} (citation domain) and \textbf{History} (co-purchase domain) are used to fine-tune the base LLM, enabling it to capture the semantic characteristics of two distinct relational types and to learn reasoning under relational constraints.  
To ensure that evaluation reflects genuine cross-domain and cross-relation generalisation, datasets from the hyperlink and social domains are deliberately excluded from training.  
All datasets are randomly split into 60\%/20\%/20\% for training, validation, and testing, respectively.  
Prompt templates for generative LLMs are listed in Table~\ref{tab:prompt_template}.  
Qwen2.5-7B-Instruct serves as the base model for TRN-R1-Zero.  
Low-Rank Adaptation (LoRA~\cite{lora}) is employed for memory-efficient fine-tuning, with the rank set to 64.  
All experiments are conducted on a single AMD MI300X GPU.

For the margin-gain computation, the SGC aggregator in Eq.~(\ref{eq:sgc}) is applied with $K{=}1$.
The temperature in the reshaping factor of Eq.~(\ref{eq:reshape}) is set to $\alpha{=}10$, amplifying the reward contribution of samples whose classification margin shifts substantially under neighbourhood aggregation.

\subsection{Overall Results}

The overall zero-shot node classification results are presented in Table~\ref{tab:zeroshot}. \textbf{TRN-R1-Zero attains the highest average Accuracy and Macro-F1 across all datasets}, validating its effectiveness and superior generalisation across domains.

LLaGA exhibits noticeably lower performance, indicating limited domain transferability, as its mapping layer to align graph embeddings with LLM embeddings is trained on a source graph and struggles to generalise to unseen graphs.  
In contrast, ZeroG achieves competitive results, since its post-encoding information aggregation does not compromise generalisation ability.

Among pure LLMs, GPT-4o achieves the best performance on Cora and Photo, whereas Qwen2.5-14B-Instruct surpasses GPT-4o on WikiCS and Instagram.  
For models of comparable scale, Qwen2.5-7B-Instruct consistently outperforms LLaMA-3.1-8B across all datasets.  
Although the smaller Qwen2.5-1.5B-Instruct exceeds LLaMA-3.1-8B on WikiCS and Photo, it falls behind on Cora.  
These results suggest that both larger model capacity and instruction tuning contribute positively to zero-shot relational reasoning.

For reasoning-based LLMs, TRN-R1-Zero not only achieves the best or near-best results overall but also demonstrates strong generalisation capability.  
Despite being trained only on the Citeseer and History datasets and without any exposure to test graphs, TRN-R1-Zero performs well on both in-domain datasets (Cora, Photo) and out-of-domain datasets (WikiCS, Instagram).  

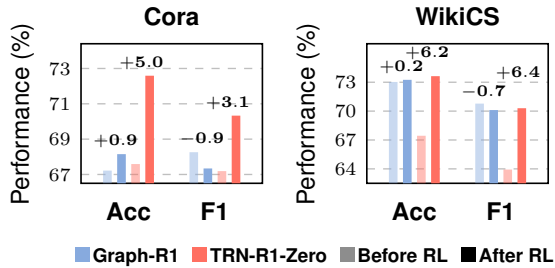
\begin{figure}[!h]
    \centering
    \resizebox{\linewidth}{!}{\input{figures/rl_gain_combined.tex}}
    \caption{Performance comparison with RL training between our TRN-R1-Zero (\textcolor{red}{red}) and Graph-R1 (\textcolor{blue}{blue}).}
    \label{fig:rl_perf_gain}
\end{figure}

\subsection{Generalisation to Graph and Edge Tasks}
\label{sec:exp-cross-task}
Although TRN-R1-Zero is trained only on node classification, its zero-shot ability is further examined on two unseen tasks across Expla-Graph, WikiCS-Link and Instagram-Link.
For graph-level reasoning, TRN-R1-Zero improves over the base model at both 7B and 14B scales, and under the same Qwen2.5-14B backbone, it also surpasses Graph-R1, even though Graph-R1 is explicitly trained on graph-level tasks. 
For edge-level prediction, the gains here are particularly substantial (e.g., +16.10 on WikiCS-Link for 7B), indicating effective transfer of relational reasoning, and again TRN-R1-Zero exceeds Graph-R1 under the same 14B backbone despite never being trained on edge-level supervision. \textbf{Together, these results highlight the zero-shot ability of TRN-R1-Zero on graph and edge level tasks, derived from the effective neighbour-aware reasoning training only on node level tasks.}

\begin{table}[t]
\centering
\small
\setlength{\tabcolsep}{3pt}
\resizebox{\columnwidth}{!}{
\renewcommand{\arraystretch}{1.1}
\begin{tabular}{@{}c l c c c@{}}
\toprule
\textbf{Scale} & \textbf{Model} & \textbf{Expla-Graph} & \textbf{WikiCS-Link} & \textbf{Insta-Link} \\
\midrule
\multirow{2}{*}{\rotatebox[origin=c]{90}{\textbf{7B}}}
 & Qwen2.5 & 84.12 & 52.10 & 64.90 \\
 & \cellcolor{rowShade}\textbf{\textsc{TRN-R1-Zero}}
& \cellcolor{rowShade}87.18\,{\scriptsize\textcolor{deltaColor}{+3.06}}
& \cellcolor{rowShade}68.20\,{\scriptsize\textcolor{deltaColor}{+16.10}}
& \cellcolor{rowShade}66.80\,{\scriptsize\textcolor{deltaColor}{+1.90}} \\
\cmidrule(lr){1-5}
\multirow{3}{*}{\rotatebox[origin=c]{90}{\textbf{14B}}}
 & Qwen2.5 & 89.89 & 72.10 & 71.80 \\
 & Graph-R1    & 89.71 & 48.90 & 56.40 \\
 & \cellcolor{rowShade}\textbf{\textsc{TRN-R1-Zero}}
& \cellcolor{rowShade}90.25\,{\scriptsize\textcolor{deltaColor}{+0.36}}
& \cellcolor{rowShade}73.90\,{\scriptsize\textcolor{deltaColor}{+1.80}}
& \cellcolor{rowShade}74.20\,{\scriptsize\textcolor{deltaColor}{+2.40}} \\
\bottomrule
\end{tabular}
}
\caption{Zero-shot performance on graph reasoning and link prediction. Trained only on node level tasks.}
\label{tab:task_generalisation}
\end{table}

\subsection{Effectiveness of Neighbour-aware RL}

This experiment investigates how our proposed neighbour-aware RL post-training can enhance the zero-shot node classification capability of LLMs.  
The comparison is between using the vanilla CoT distillation in Graph-R1 (14B) and our TRN-R1-Zero (7B). The evaluation is conducted on the Cora and WikiCS datasets, which represent the citation and hyperlink domains, using accuracy and macro-F1 as the primary metrics.  

Figure~\ref{fig:rl_perf_gain} shows the performance gains achieved through RL training on the Cora and WikiCS datasets.  
TRN-R1-Zero consistently yields larger improvements in both Accuracy and F1, whereas Graph-R1 even experiences a decline in F1 across both datasets.  
These results indicate that the neighbour-aware reward design effectively stabilises the optimisation process and promotes more balanced metric improvements. \textbf{TRN-R1-Zero not only delivers consistent and robust performance gains through reinforcement learning, but also exhibits superior optimisation stability and generalisation} compared with baselines.

\subsection{Effectiveness of Margin Gain}
\begin{figure}[!h]
    \centering
    \begin{subfigure}[b]{0.48\linewidth}
        \centering
        \includegraphics[width=\linewidth]{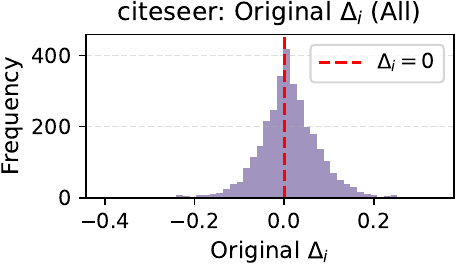}
    \end{subfigure}
    \hfill
    \begin{subfigure}[b]{0.48\linewidth}
        \centering
        \includegraphics[width=\linewidth]{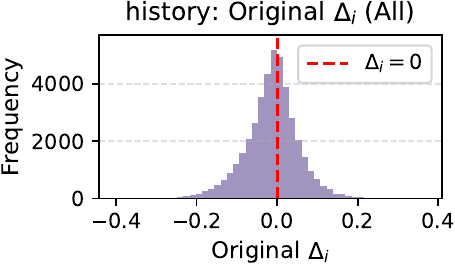}
    \end{subfigure}
    \caption{Original margin gain values $\Delta_i$ across the training datasets (Citeseer and History). These results demonstrate the distribution of impact from neighbour information towards the target node, motivating the neighbour-aware reward design.}
    \label{fig:delta}
\end{figure}

The margin gain visualisations in Figure~\ref{fig:delta} provide an intuitive view of how neighbour aggregation influences decision confidence during RL training. 
Specifically, a positive $\Delta_i$ indicates that neighbour aggregation shifts the target embedding closer to the ground-truth label embedding, whereas a negative $\Delta_i$ indicates the opposite effect. A larger $|\Delta_i|$ therefore reflects a stronger influence of neighbour information on the classification of the target node, and such high-impact samples are the ones most worth emphasising during policy optimisation.

\begin{figure}[!h]
    \centering
    \begin{subfigure}[t]{0.48\linewidth}
        \centering
        \includegraphics[width=\linewidth]{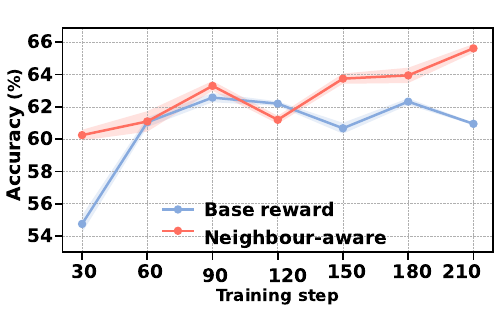}
        \caption{Cora Evaluation}
        \label{fig:margin_ablation}
    \end{subfigure}
    \hfill
    \begin{subfigure}[t]{0.48\linewidth}
        \centering
        \includegraphics[width=\linewidth]{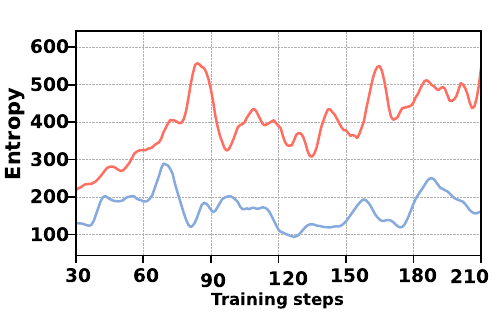}
        \caption{Entropy}
        \label{fig:entropy}
    \end{subfigure}

    \vspace{0.8em}

    \begin{subfigure}[t]{0.48\linewidth}
        \centering
        \includegraphics[width=\linewidth]{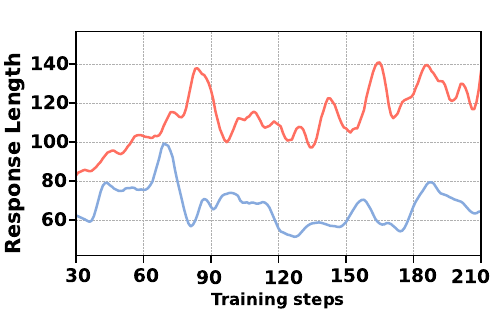}
        \caption{Response Length}
        \label{fig:length}
    \end{subfigure}
    \hfill
    \begin{subfigure}[t]{0.48\linewidth}
        \centering
        \includegraphics[width=\linewidth]{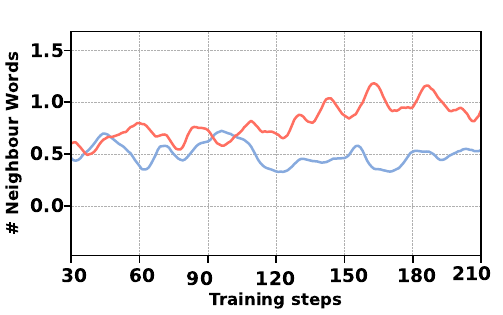}
        \caption{\#Neighbour Word}
        \label{fig:count}
    \end{subfigure}

    \caption{Accuracy comparison between base reward and neighbour-aware reward across Cora dataset.  
    Neighbour-aware shaping consistently improves both optimisation stability and reasoning depth.}
\end{figure}

To examine the effect of the neighbour-aware policy objective on training dynamics, the Qwen2.5-1.5B-Instruct model is used as the base policy model for computational efficiency.  
Two reward variants are compared: (i) the base reward without scaling and (ii) the $\exp(|\Delta_i|)$-scaled reward.  
The Cora dataset is used for evaluation. Figure~\ref{fig:margin_ablation} illustrates the average accuracy across training steps under both settings.  
The results show that incorporating neighbour-aware reward shaping stabilises optimisation and yields more consistent performance improvements compared with the unshaped baseline.  
Each checkpoint model is evaluated five times on the Cora dataset to ensure robustness.
The training statistics in Figure~\ref{fig:entropy}, \subref{fig:length}, and \subref{fig:count} further support the effectiveness of neighbour-aware shaping:  
the entropy remains relatively high, encouraging the policy model to explore more diverse and optimised responses rather than over-exploit early patterns;  
meanwhile, the average response length steadily increases, suggesting that the model performs deeper reasoning.  
Additionally, the rising frequency of the word “neighbour” indicates that the model gradually learns to leverage relational context more effectively. Our \textbf{neighbour-aware margin gain enhances both the stability and utilisation of neighbourhood in reasoning depth} for node classification.

\subsection{Impact of Different LLM Backbones}
To assess the generality of \textsc{TRN-R1-Zero} across LLM backbones, models spanning different families and scales are trained, including Llama-3.2-3B, Llama-3.1-8B, and Qwen2.5-14B. Across all families and scales, \textsc{TRN-R1-Zero} consistently improves zero-shot node classification, with the largest gains observed on smaller backbones (e.g., +14.4 and +9.0 in average accuracy on the 3B and 8B models, respectively). These results indicate that the \textbf{proposed training paradigm is not tied to a specific backbone and generalises across architectures and scales.}

\begin{figure}[t]
\centering
\input{figures/llm_scaling.tex}
\caption{Zero-shot node classification accuracy across three model scales.
Green annotations show absolute gain over the base LLM.}
\label{fig:scaling}
\end{figure}
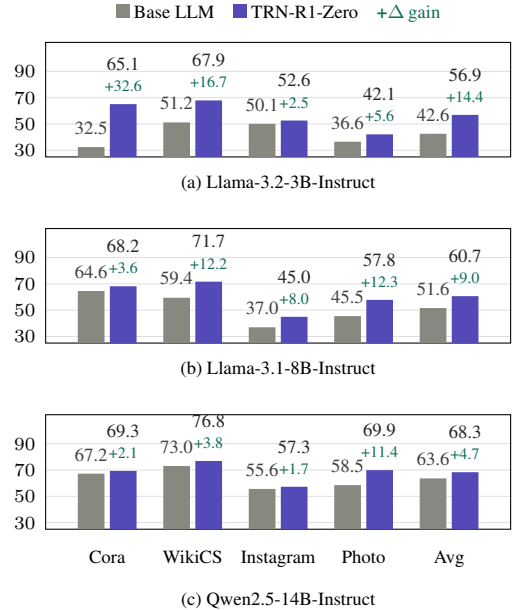

\subsection{Performance under Supervised Settings}

In the supervised setting, 60\% of each dataset is used for training and 20\% for testing.  
This configuration follows the common practice in GNN and GNN-adapter frameworks (e.g., LLaGA) for LLMs.  
Since direct supervision signals are available from the training-testing split within the same dataset, traditional supervised models such as Graph Convolutional Networks (GCNs) can perform effectively.  
As shown in Table~\ref{tab:sup}, TRN-R1-Zero still outperforms both GCN and LLaGA under the supervised setting, demonstrating strong optimisation and reasoning capabilities even when explicit labels are provided.  
\textbf{Overall, TRN-R1-Zero maintains superior performance and stable optimisation behaviour under supervised learning conditions.}

\begin{table}[ht]
    \centering
    \small
    \begin{tabular}{c|c|c}
    \toprule
         Model & Citeseer & History \\
        \midrule
        GCN & 76.45 & 84.23 \\
        LLaGA & 76.73 & 85.56  \\
        \midrule
        TRN-R1-Zero & \textbf{77.74} & \textbf{86.71} \\
    \bottomrule
    \end{tabular}
    \caption{
    Comparison of TRN-R1-Zero with GCN~\cite{gcn} and LLaGA under the supervised setting on the Citeseer and History datasets.
    }
    \label{tab:sup}
\end{table}

\subsection{Case Study and Efficiency}

Compared with other reasoning-based methods, \textbf{TRN-R1-Zero enjoys a much higher efficiency by (1) employing a substantially smaller model; and (2) significantly shortens the average response length, producing more concise yet effective reasoning traces} while achieving superior performance (see Box2 and Box3).  
This improvement not only reduces GPU memory consumption but also accelerates inference, demonstrating that reinforcement learning alone can promote efficient reasoning behaviour without relying on large-scale supervised fine-tuning or external LLMs.

\begin{table}[!h]
\centering
\small
\begin{tabular}{c|c|c|c|c}
\toprule
Dataset & Model & Size & Resp. Len. & Acc. \\
\midrule
Cora & TRN-R1-Zero & 7B & 152.57 & 72.59 \\
Cora & Graph-R1 & 14B & 930.47 & 68.15 \\
\midrule
WikiCS & TRN-R1-Zero & 7B & 129.27 & 73.63 \\
WikiCS & Graph-R1 & 14B & 1129.22 & 73.25 \\
\bottomrule
\end{tabular}
\caption{
Comparison of model size, average response length (Resp. Len.), and accuracy (Acc.) between TRN-R1-Zero and Graph-R1 on Cora and WikiCS.
}
\label{tab:efficiency}
\end{table}

\begin{center}
    \begin{response}[title={Box 2: Response of TRN-R1-Zero}, label=prompt:response1]
    \noindent \# Response for Node 11 in Cora by TRN-R1-Zero\\ \\
    \texttt{<think>
    The target node discusses the MIN-FEATURES bias in machine learning, which is related to the selection of relevant features. It talks about learning algorithms, sample complexity, and generalization performance. The focus is on understanding and improving the learning process by considering the relevance of features, which is a theoretical aspect of machine learning.
    \
    \textcolor{orange}{Looking at the neighbor nodes}:
    - One discusses maximizing coverage in learning algorithms.
    - Another focuses on prototype and feature selection using sampling and random mutation hill climbing.
    - The third uses compression-based methods for feature subset selection.
    \
    \textcolor{orange}{All these topics are closely related to feature selection and learning algorithms, which fall under the broader category of machine learning theory}. Therefore, the most appropriate category for the target node would be "Theory" (category 4).
    </think>
    <answer>
    4~\ding{51}
    </answer>}
    \label{fig:reasoning}
\end{response}
\end{center}

\begin{center}
    \begin{response}[title={Box 3: Response of Graph-R1}, label=prompt:response2]
    \noindent \# Response for Node 11 in Cora by Graph-R1\\ \\
    \texttt{Okay, let's try to figure out how to classify the paper segment represented by node0. ...\textbf{38 words}...
    \
    First, I need to understand what each node is about. Node0's paper is about the MIN-FEATURES bias in machine learning, focusing on feature selection. ...\textbf{35 words}...
    \
    Looking at the connections: node0 is connected to node1, node2, and node3. Node1 talks about feature subset selection using Information Theory. Node2 discusses MDL (Minimum Description Length) for feature selection. Node3 is about constructive induction and feature selection with a non-greedy strategy.
    \
    ...\textbf{334 words}...
    Given the options, \textcolor{orange}{"Rule\_Learning" (0) seems the most fitting because it directly relates to the algorithms discussed (ID3, FOCUS) and the context of feature selection in decision trees. "Theory" is a possible contender but might be too broad. The other categories don't align as well}. 
    \
    Therefore, the best fit is Rule\_Learning (0).
    </think>
    \
    <structure>
    The subgraph forms a star topology with node0 at the center connected to nodes1,2,3. ...\textbf{43 words}...
    </structure>
    \
    <semantic>
   ...\textbf{40 words}...\textcolor{orange}{"Theory" (category4) is plausible due to PAC-learning and Information Theory, but the emphasis on practical algorithms (ID3, FOCUS) leans more toward applied methods.} "Probabilistic\_Methods" (6) is less fitting as MDL is information-theoretic, not probabilistic. Other categories (Neural\_Networks, Case\_Based, Genetic\_Algorithms) are irrelevant.
    </semantic>
    \
    <comprehensive>
    Candidate answers:
    1. **Rule\_Learning (0)**: Directly supported by ID3/FOCUS comparisons, feature selection in decision trees, and algorithmic focus.
    2. **Theory (4)**: PAC-learning bounds and Information Theory overlap, but less emphasis on theoretical frameworks compared to applied algorithms.
    3. **Probabilistic\_Methods (6)**: MDL relates to information theory but lacks explicit probabilistic modeling.
    \
    Rethinking:
    - \textcolor{orange}{**Rule\_Learning** best captures the practical algorithm comparisons (ID3, FOCUS) and feature selection in decision trees.}
    - **Theory** is secondary but insufficient due to the applied focus of node0.
    - Other categories are invalid based on content exclusion.
    \
    Final Answer: 0 (Rule\_Learning)~\ding{55}}
\end{response}
\end{center}

\section{Conclusion}
This paper introduces TRN-R1-Zero, a reinforcement learning-only framework for text-rich node reasoning. A neighbour-aware group relative policy optimisation objective models the impact of neighbouring information during training with a novel margin gain. Without distillation or supervised fine-tuning, TRN-R1-Zero directly optimises a base LLM for strong zero-shot reasoning. Extensive experiments verify the effectiveness of TRN-R1-Zero.

\section*{Limitations}

For zero-shot text-rich network (TRN) tasks, LLMs must not only extract useful information from the target node text, neighbour node texts, and candidate label texts, but also comprehend the underlying semantics of these texts.  
Therefore, if the base LLM lacks sufficient domain knowledge, reinforcement learning may offer limited improvements, as such knowledge primarily originates from the pre-training phase.

\section*{Acknowledgments}
This research has been supported by Australian Research Council
Discovery Projects (CE200100025, DP230101196 and DE250100919).

\bibliography{custom}
\newpage
\appendix

\section{{Text Quality for Current Text-rich Networks}}
\label{app:meaningless}

Textual nodes in text-rich networks often meet noisy or incomplete text. It is hard for a LLM directly give a reasonable label for such nodes. Examples for Cora and Photo datasets are shown in Table~\ref{tab:noise_examples}.

\begin{table}[ht]
    \centering
    \begin{tabular}{p{0.1\linewidth}|p{0.8\linewidth}}
        \hline
        Cora & TABLE DES MATI ERES 1 Apprentissage et approximation les techniques de regularisation 3 1.1 Introduction: \\
        \hline
        Photo & Good product, good price good price without shipping fee. With shipping fee, it is still a good deal. \\
        \hline
    \end{tabular}
    \caption{Examples of meaningless and noisy texts in TRNs.}
    \label{tab:noise_examples}
\end{table}

\section{Prompt Design for Edge-level and Graph-level Tasks}
\label{app:prompts}

The following prompts are used to evaluate TRN-R1-Zero for edge-level and graph-level tasks. The differences between them are the graph prompt and the task instruction parts.

\begin{table*}[ht]
  \centering
  \renewcommand{\arraystretch}{1.25}
  \begin{tabularx}{\textwidth}{
      >{\centering\arraybackslash}m{0.13\textwidth} 
      m{0.14\textwidth} 
      m{0.18\textwidth} 
      X
  }
    \toprule
    \textbf{Relation} & \textbf{Dataset} & \textbf{Node Type} & \textbf{Labels} \\
    \midrule
    \multirow{2}{*}{\textbf{Citation}}
      & Cora     & Paper segment & Rule\_Learning; Neural\_Networks; Case\_Based; Genetic\_Algorithms; Theory; Reinforcement\_Learning; Probabilistic\_Methods \\
      & Citeseer & Paper segment & Agents; ML (Machine Learning); IR (Information Retrieval); DB (Databases); HCI (Human-Computer Interaction); AI (Artificial Intelligence) \\
    \midrule
    \multirow{1}{*}{\textbf{Hyperlink}}
      & WikiCS     & Wikipedia article & Computational Linguistics; Databases; Operating Systems; Computer Architecture; Computer Security; Internet Protocols; Computer File Systems; Distributed Computing Architecture; Web Technology; Programming Language Topics \\
    \midrule
    \multirow{1}{*}{\textbf{Social}}
      & Instagram     & Instagram User Bio & Normal Users; Commercial Users \\
    \midrule
    \multirow{2}{*}{\textbf{Co-purchase}}
      & History  & Product description & World; Americas; Asia; Military; Europe; Russia; Africa; Ancient Civilizations; Middle East; Historical Study \& Educational Resources; Australia \& Oceania; Arctic \& Antarctica \\
      & Photo    & Customer review     & Video Surveillance; Accessories; Binoculars \& Scopes; Video; Lighting \& Studio; Bags \& Cases; Tripods \& Monopods; Flashes; Digital Cameras; Film Photography; Lenses; Underwater Photography \\
    \midrule
    \multirow{1}{*}{\textbf{Commonsense}}
      & Expla-Graph  & Commonsense concept & Support; Counter \\

    \bottomrule
  \end{tabularx}
  \caption{Meta information for benchmark TRNs grouped by relation type.}
  \label{tab:dataset_metadata}
\end{table*}

\begin{center}
    \begin{prompt}[title={Edge-level Prompt for TRN-R1-Zero}, label=tab:prompt_template_edge]
    \# System Prompt

    You are a helpful AI Assistant that provides well-reasoned and detailed responses.
You first think about the reasoning process as an internal monologue and then provide the user with the answer.
Respond in the following format: \\
\texttt{<think>} \\
\texttt{...} \\
\texttt{</think>} \\
\texttt{<answer>} \\
\texttt{...} \\
\texttt{</answer>} \\

\# Graph Prompt

Source node: \texttt{\{source\_node\}} \\
Target node: \texttt{\{target\_node\}} \\
Neighbours of source node: \texttt{\{source\_neighbors\}} \\
Neighbours of target node: \texttt{\{target\_neighbors\}} \\

\# Task Instruction

Your task is to predict whether a link exists between two nodes in a graph.
Each node represents a \texttt{\{node\_type\}}. The relation type in this graph
is \texttt{\{relation\}}. \\
Question: Based on the attributes and neighbourhood structure of both nodes,
predict whether a \texttt{\{relation\}} link exists between the source and
target nodes. Answer with 0 (no link) or 1 (link exists).
\end{prompt}
\end{center}

\begin{center}
    \begin{prompt}[title={Graph-level Prompt for TRN-R1-Zero}, label=tab:prompt_template_graph]
    \# System Prompt

    You are a helpful AI Assistant that provides well-reasoned and detailed responses.
You first think about the reasoning process as an internal monologue and then provide the user with the answer.
Respond in the following format: \\
\texttt{<think>} \\
\texttt{...} \\
\texttt{</think>} \\
\texttt{<answer>} \\
\texttt{...} \\
\texttt{</answer>} \\

\# Graph Prompt

Nodes: \texttt{\{node\_list\}} \\
Relationships: \texttt{\{edge\_list\}} \\

\# Task Instruction

Your task is to determine if two arguments support or counter each other,
based on the provided commonsense graph. The commonsense graph is defined by
nodes and their relationships. \\
Based on this graph, consider the following: \texttt{\{question\}}. \\
Your answer must be a single integer ID, where 0 means support and 1 means
counter.
\end{prompt}
\end{center}

\section{Extended Dataset Statistics}
The following tables include the extended dataset statistics (Table~\ref{tab:datasets}) with detailed description of meta data like label in text for each dataset (Table~\ref{tab:dataset_metadata}).

\begin{table*}[ht]
  \centering
  \renewcommand{\arraystretch}{1.15}
  \begin{tabular*}{\textwidth}{@{\extracolsep{\fill}} c c c c c c c C{1.2cm} }
    \toprule
    \textbf{Domain} & \textbf{Dataset} & \textbf{\#Nodes} & \textbf{\#Graphs} & \textbf{\#Edges}
    & \textbf{Avg. Deg.} & \textbf{Homo.} & \textbf{\#Classes} \\
    \midrule
    \multirow{2}{*}{Citation}
      & Citeseer$^{*}$ & 3,186 & 1 & 8,450   & 2.65 & 0.72 & 6  \\
      & Cora     & 2,708 & 1 & 5,429   & 3.90 & 0.83 & 7  \\
    \midrule
    \multirow{1}{*}{Webpage}
      & WikiCS     & 11,701 & 1 & 431,206 & 36.85 & 0.68 & 10 \\
    \midrule
    \multirow{1}{*}{Social}
    & Instagram  & 11,339 & 1 & 144,010  & 12.70 & 0.59 & 2 \\
    \midrule
    \multirow{2}{*}{Co-purchase}
      & Photo    & 48,362 & 1 & 873,782  & 18.07 & 0.79 & 12 \\
      & History$^{*}$  & 41,551 & 1 & 503,180  & 12.11 & 0.78 & 12 \\
    \midrule
    \multirow{1}{*}{Commonsense}
      & Expla-Graph    & 5.2 & 2,766 & 4.3  & - & - & 2 \\
    \bottomrule
  \end{tabular*}
  \caption{Statistics of benchmark datasets. 
  Datasets marked with $^{*}$ (\textit{Citeseer} and \textit{History}) are used for RL training, while the others are held out for evaluation and generalisation studies.}
  \label{tab:datasets}
\end{table*}

\end{document}

%% file: figures/trn.tex
\usetikzlibrary{calc,arrows.meta}

\providecommand{\defineTRNcolor}[2]{\definecolor{#1}{HTML}{#2}}
\defineTRNcolor{citeBg}{ECEEF5}   \defineTRNcolor{citeAc}{3B4A78}  \defineTRNcolor{citeHl}{CED5E6}
\defineTRNcolor{hypBg}{F3E9E4}    \defineTRNcolor{hypAc}{94513E}   \defineTRNcolor{hypHl}{E5CCC0}
\defineTRNcolor{socBg}{EEF1E8}    \defineTRNcolor{socAc}{566E43}   \defineTRNcolor{socHl}{D3DDC4}
\defineTRNcolor{shopBg}{F4ECDB}   \defineTRNcolor{shopAc}{8D6A28}  \defineTRNcolor{shopHl}{E8D6A8}
\defineTRNcolor{taskBg}{F5F3EE}   \defineTRNcolor{taskBd}{CFCCC4}  \defineTRNcolor{taskChip}{2D2D2A}
\defineTRNcolor{iconStroke}{5A5A57} %
\defineTRNcolor{iconFaint}{8F8F8C}  %

\providecommand{\paperinline}[2]{%
  {\colorlet{ppstroke}{#1}%
   \tikz[baseline=-0.25ex, x=1mm, y=1mm]{%
     \draw[line width=0.4pt, draw=ppstroke, fill=#2, rounded corners=0.15]
       (0,0) rectangle (3.2, 4.1);
     \fill[ppstroke!85] (3.2, 4.1) -- (3.2, 3.2) -- (2.3, 4.1) -- cycle;
     \draw[line width=0.25pt, ppstroke!65]
       (0.55, 2.55) -- (2.65, 2.55)
       (0.55, 1.65) -- (2.65, 1.65)
       (0.55, 0.75) -- (2.10, 0.75);
   }}%
}

\resizebox{\columnwidth}{!}{%
\begin{tikzpicture}[
    panel/.style  = {rounded corners=2.5pt, minimum width=2.15cm, minimum height=2.35cm, inner sep=2pt},
    ptitle/.style = {font=\sffamily\bfseries, anchor=north, inner sep=1.5pt},
    doc/.style    = {draw=iconStroke, line width=0.6pt, fill=white, rectangle,
                     minimum width=5mm, minimum height=6mm, inner sep=0pt, rounded corners=0.3pt},
    win/.style    = {draw=iconStroke, line width=0.6pt, fill=white, rectangle,
                     minimum width=7mm, minimum height=5mm, inner sep=0pt, rounded corners=0.5pt},
    nodec/.style  = {draw=iconStroke, line width=0.6pt, fill=white, circle,
                     minimum size=5mm, inner sep=0pt},
    arr/.style    = {-{Latex[length=1.5mm,width=1.5mm]}, line width=0.6pt},
    darr/.style   = {{Latex[length=1.3mm,width=1.3mm]}-{Latex[length=1.3mm,width=1.3mm]}, line width=0.6pt},
    dashedge/.style = {line width=0.6pt, dash pattern=on 1.2pt off 1.2pt},
    edge/.style   = {line width=0.6pt},
  ]

  \def\panelstep{2.35}

  \begin{scope}[local bounding box=P1]
    \node[panel, fill=citeBg] (p1) at (0,0) {};
    \node[ptitle, text=citeAc] at (p1.north) {Citation};
    \node[doc] (c1) at ($(p1.center)+(-0.55,0.20)$) {};
    \node[doc] (c2) at ($(p1.center)+(0.55,0.20)$) {};
    \node[doc, draw=citeAc, fill=citeHl, line width=0.8pt] (c0)
          at ($(p1.center)+(0,-0.55)$) {};
    \foreach \n in {c1,c2}{
      \fill[iconStroke!85]
        ($(\n.north east)+(0,-1.2mm)$) -- (\n.north east) -- ($(\n.north east)+(-1.2mm,0)$) -- cycle;
      \draw[line width=0.3pt, iconFaint]
        ($(\n.west)+(0.9mm, 1.3mm)$) -- ($(\n.east)+(-0.9mm, 1.3mm)$)
        ($(\n.west)+(0.9mm, 0.2mm)$) -- ($(\n.east)+(-0.9mm, 0.2mm)$)
        ($(\n.west)+(0.9mm,-0.9mm)$) -- ($(\n.east)+(-0.9mm,-0.9mm)$);
    }
    \fill[citeAc!85]
      ($(c0.north east)+(0,-1.2mm)$) -- (c0.north east) -- ($(c0.north east)+(-1.2mm,0)$) -- cycle;
    \draw[line width=0.3pt, citeAc!55]
      ($(c0.west)+(0.9mm, 1.3mm)$) -- ($(c0.east)+(-0.9mm, 1.3mm)$)
      ($(c0.west)+(0.9mm, 0.2mm)$) -- ($(c0.east)+(-0.9mm, 0.2mm)$)
      ($(c0.west)+(0.9mm,-0.9mm)$) -- ($(c0.east)+(-0.9mm,-0.9mm)$);
    \draw[arr, citeAc] (c0.north) -- ($(c1.south)+(0.1,0)$);
    \draw[arr, citeAc] (c0.north) -- ($(c2.south)+(-0.1,0)$);
  \end{scope}

  \begin{scope}[local bounding box=P2, shift={(\panelstep,0)}]
    \node[panel, fill=hypBg] (p2) at (0,0) {};
    \node[ptitle, text=hypAc] at (p2.north) {Hyperlink};
    \node[win] (w1) at ($(p2.center)+(-0.55,0.25)$) {};
    \node[win] (w2) at ($(p2.center)+(0.55,0.25)$) {};
    \node[win, draw=hypAc, fill=hypHl, line width=0.8pt] (w0)
          at ($(p2.center)+(0,-0.50)$) {};
    \foreach \n in {w1,w2,w0}{
      \path[fill=black!10, draw=none]
        ($(\n.north west)+(0.3mm,-0.3mm)$) rectangle ($(\n.north east)+(-0.3mm,-1.3mm)$);
      \foreach \i/\c in {0/iconFaint, 1/iconFaint, 2/iconFaint}{
        \fill[\c] ($(\n.north west)+(1mm + \i*1mm,-0.8mm)$) circle (0.18mm);
      }
      \draw[line width=0.3pt, iconFaint]
        ($(\n.west)+(0.9mm,-0.3mm)$) -- ($(\n.east)+(-0.9mm,-0.3mm)$)
        ($(\n.west)+(0.9mm,-1.2mm)$) -- ($(\n.east)+(-2.5mm,-1.2mm)$);
    }
    \draw[darr, hypAc] (w0.north) to[bend right=25] (w1.south);
    \draw[darr, hypAc] (w0.north) to[bend left=25] (w2.south);
  \end{scope}

  \begin{scope}[local bounding box=P3, shift={(2*\panelstep,0)}]
    \node[panel, fill=socBg] (p3) at (0,0) {};
    \node[ptitle, text=socAc] at (p3.north) {Social Link};
    \node[nodec, draw=socAc, fill=socHl, line width=0.8pt] (s0)
          at ($(p3.center)+(0,-0.05)$) {};
    \foreach \ang/\name in {55/n1, 125/n2, 195/n3, 270/n4, 340/n5}{
      \node[nodec] (\name) at ($(s0.center)+(\ang:0.75)$) {};
    }
    \foreach \nbr in {n1,n2,n3,n4,n5}{
      \draw[edge, socAc] (s0) -- (\nbr);
    }
    \foreach \n in {s0,n1,n2,n3,n4,n5}{
      \fill[iconStroke] ($(\n.center)+(0,0.7mm)$) circle (0.35mm);
      \draw[line width=0.35pt, iconStroke]
        ($(\n.center)+(-0.9mm,-0.5mm)$) .. controls
        ($(\n.center)+(-0.9mm,0.0mm)$) and ($(\n.center)+(0.9mm,0.0mm)$) ..
        ($(\n.center)+(0.9mm,-0.5mm)$);
    }
  \end{scope}

  \begin{scope}[local bounding box=P4, shift={(3*\panelstep,0)}]
    \node[panel, fill=shopBg] (p4) at (0,0) {};
    \node[ptitle, text=shopAc] at (p4.north) {Co-purchase};
    \tikzset{
      pics/bag/.style args={#1}{
        code={
          \draw[line width=0.5pt, shopAc]
            (-0.9mm, 1.8mm) arc (180:0:0.9mm);
          \draw[line width=0.6pt, draw=iconStroke, fill=#1]
            (-1.6mm,-2.2mm) -- (-2mm,1.8mm) -- (2mm,1.8mm) -- (1.6mm,-2.2mm) -- cycle;
        }
      }
    }
    \node (b1)  at ($(p4.center)+(-0.55,0.20)$) {}; \path (b1) pic{bag=white};
    \node (b2)  at ($(p4.center)+(0.55,0.20)$)  {}; \path (b2) pic{bag=white};
    \node (b0)  at ($(p4.center)+(0,-0.50)$)    {}; \path (b0) pic{bag=shopHl};
    \draw[dashedge, shopAc] ($(b0)+(-0.2,0.18)$) -- ($(b1)+(0.16,-0.22)$);
    \draw[dashedge, shopAc] ($(b0)+(0.2,0.18)$)  -- ($(b2)+(-0.16,-0.22)$);
  \end{scope}

  \coordinate (taskL) at ($(P1.west |- 0,0)+(0,-2.15)$);
  \coordinate (taskR) at ($(P4.east |- 0,0)+(0,-2.15)$);
  \node[draw=taskBd, line width=0.45pt, rounded corners=2.5pt,
        fill=taskBg,
        minimum width=9.20cm,
        minimum height=1.70cm, anchor=center,
        inner xsep=6pt, inner ysep=5pt]
        (task) at ($(taskL)!0.5!(taskR)$) {};

  \node[anchor=north west, fill=taskChip, text=white, rounded corners=1.5pt,
        inner xsep=3pt, inner ysep=1.5pt,
        font=\sffamily\bfseries\fontsize{6.5}{7.5}\selectfont]
        (chip) at ($(task.north west)+(5pt,-5pt)$) {TASK};
  \node[anchor=west, font=\sffamily\small, text=black!55]
        at (chip.east)
        {\hspace{4pt}Zero-shot node classification on TRNs};

  \node[anchor=north west, align=left, text width=8.80cm,
        font=\itshape, text=black!80, inner sep=0pt]
        at ($(task.north west)+(6pt,-17pt)$)
        {Given the target paper\,%
         \paperinline{citeAc}{citeHl}\,%
         and its cited papers\,%
         \paperinline{iconStroke}{white}\,%
         and\,%
         \paperinline{iconStroke}{white}\,,
         what is the most suitable category?};
\end{tikzpicture}%
}

%% file: figures/rl_gain_combined.tex
\definecolor{graphRoneColor}{HTML}{87AADE} %
\definecolor{trnRoneColor}  {HTML}{FF6F61} %

\begin{tikzpicture}[
    deltalabel/.style={font=\tiny\sffamily\bfseries, inner sep=1pt, anchor=south},
    legendsq/.style={draw=none},
]
\pgfplotsset{
    rlpanel/.style={
        ybar,
        bar width=0.10,
        width=0.51\linewidth,
        height=3.4cm,
        xmin=0.45, xmax=2.55,
        xtick={1,2},
        xticklabels={Acc, F1},
        xticklabel style={font=\small\sffamily\bfseries, yshift=1pt},
        yticklabel style={font=\scriptsize\sffamily\bfseries},
        ylabel style={font=\small\sffamily, yshift=-2pt},
        title style={font=\small\sffamily\bfseries, yshift=-3pt},
        ymajorgrids=true,
        grid style={dashed, gray!50, line width=0.5pt},
        major tick style={draw=none},
        minor tick style={draw=none},
        every axis plot/.append style={draw=none},
        axis line style={line width=0.6pt},
        clip=false,
    },
}

\begin{groupplot}[
    group style={
        group size=2 by 1,
        horizontal sep=1.35cm,
    },
    rlpanel,
]

\nextgroupplot[
    title={Cora},
    ylabel={Performance (\%)},
    ymin=66.5, ymax=74.4,
    ytick={67,69,71,73},
]
\addplot[fill=graphRoneColor, fill opacity=0.45, draw=none] coordinates {(1,67.22) (2,68.26)};
\addplot[fill=graphRoneColor, fill opacity=1.00, draw=none] coordinates {(1,68.15) (2,67.34)};
\addplot[fill=trnRoneColor,   fill opacity=0.45, draw=none] coordinates {(1,67.59) (2,67.19)};
\addplot[fill=trnRoneColor,   fill opacity=1.00, draw=none] coordinates {(1,72.59) (2,70.33)};

\node[deltalabel] at (axis cs:0.85, 68.35) {$\mathbf{+0.9}$};
\node[deltalabel] at (axis cs:1.15, 72.79) {$\mathbf{+5.0}$};
\node[deltalabel] at (axis cs:1.85, 68.46) {$\mathbf{-0.9}$};
\node[deltalabel] at (axis cs:2.15, 70.53) {$\mathbf{+3.1}$};

\nextgroupplot[
    title={WikiCS},
    ylabel={Performance (\%)},
    ymin=62.5, ymax=77.0,
    ytick={64,67,70,73},
]
\addplot[fill=graphRoneColor, fill opacity=0.45, draw=none] coordinates {(1,73.03) (2,70.78)};
\addplot[fill=graphRoneColor, fill opacity=1.00, draw=none] coordinates {(1,73.25) (2,70.11)};
\addplot[fill=trnRoneColor,   fill opacity=0.45, draw=none] coordinates {(1,67.44) (2,63.93)};
\addplot[fill=trnRoneColor,   fill opacity=1.00, draw=none] coordinates {(1,73.63) (2,70.30)};

\node[deltalabel] at (axis cs:0.85, 73.55) {$\mathbf{+0.2}$};
\node[deltalabel] at (axis cs:1.15, 75.10) {$\mathbf{+6.2}$};
\node[deltalabel] at (axis cs:1.80, 70.95) {$\mathbf{-0.7}$};
\node[deltalabel] at (axis cs:2.20, 73.00) {$\mathbf{+6.4}$};

\end{groupplot}

\node[anchor=north, font=\scriptsize\sffamily\bfseries, yshift=-20pt]
  at ($(group c1r1.south)!0.5!(group c2r1.south)$) {%
    \tikz\draw[legendsq,fill=graphRoneColor] (0,0) rectangle (1.4ex,1.4ex);\,Graph-R1\hspace{0.7em}%
    \tikz\draw[legendsq,fill=trnRoneColor]   (0,0) rectangle (1.4ex,1.4ex);\,TRN-R1-Zero\hspace{0.7em}%
    \tikz\draw[legendsq,fill=black,fill opacity=0.45] (0,0) rectangle (1.4ex,1.4ex);\,Before RL\hspace{0.7em}%
    \tikz\draw[legendsq,fill=black] (0,0) rectangle (1.4ex,1.4ex);\,After RL%
};
\end{tikzpicture}

%% file: figures/llm_scaling.tex
\begin{tikzpicture}
\begin{groupplot}[
    group style={
        group size=1 by 3,
        vertical sep=0.95cm,
        x descriptions at=edge bottom,
    },
    width=\linewidth,
    height=3.1cm,
    ybar,
    /pgf/bar width=10pt,
    enlarge x limits=0.18,
    ymin=25, ymax=112,
    ytick={30,50,70,90},
    yticklabel style={font=\scriptsize},
    xticklabel style={font=\scriptsize},
    symbolic x coords={Cora, WikiCS, Instagram, Photo, Avg},
    xtick=data,
    ymajorgrids=true,
    grid style={very thin, gray!25},
    tick style={draw=none},
    every axis plot/.append style={draw=none},
    every node near coord/.append style={
        font=\scriptsize, align=center, yshift=1pt,
    },
]

\nextgroupplot
\addplot[fill=baseColor,
    nodes near coords, point meta=explicit symbolic,
    every node near coord/.append style={color=black!75},
] coordinates {
    (Cora,32.49)      [32.5]
    (WikiCS,51.21)    [51.2]
    (Instagram,50.12) [50.1]
    (Photo,36.55)     [36.6]
    (Avg,42.59)       [42.6]
};
\addplot+[fill=oursColor,
    nodes near coords, point meta=explicit symbolic,
    every node near coord/.append style={color=black!85},
    error bars/.cd, y dir=both, y explicit,
    error bar style={line width=0.4pt, black!60},
] coordinates {
    (Cora,65.10)       [65.1\\{\color{deltaColor}\tiny+32.6}]
    (WikiCS,67.90)     [67.9\\{\color{deltaColor}\tiny+16.7}]
    (Instagram,52.64)  [52.6\\{\color{deltaColor}\tiny+2.5}]
    (Photo,42.13)      [42.1\\{\color{deltaColor}\tiny+5.6}]
    (Avg,56.94)        [56.9\\{\color{deltaColor}\tiny+14.4}]
};

\nextgroupplot
\addplot[fill=baseColor,
    nodes near coords, point meta=explicit symbolic,
    every node near coord/.append style={color=black!75},
] coordinates {
    (Cora,64.55)      [64.6]
    (WikiCS,59.43)    [59.4]
    (Instagram,36.98) [37.0]
    (Photo,45.49)     [45.5]
    (Avg,51.61)       [51.6]
};
\addplot+[fill=oursColor,
    nodes near coords, point meta=explicit symbolic,
    every node near coord/.append style={color=black!85},
    error bars/.cd, y dir=both, y explicit,
    error bar style={line width=0.4pt, black!60},
] coordinates {
    (Cora,68.17)       [68.2\\{\color{deltaColor}\tiny+3.6}]
    (WikiCS,71.66)     [71.7\\{\color{deltaColor}\tiny+12.2}]
    (Instagram,44.95)  [45.0\\{\color{deltaColor}\tiny+8.0}]
    (Photo,57.83)      [57.8\\{\color{deltaColor}\tiny+12.3}]
    (Avg,60.65)        [60.7\\{\color{deltaColor}\tiny+9.0}]
};

\nextgroupplot
\addplot[fill=baseColor,
    nodes near coords, point meta=explicit symbolic,
    every node near coord/.append style={color=black!75},
] coordinates {
    (Cora,67.22)      [67.2]
    (WikiCS,73.03)    [73.0]
    (Instagram,55.60) [55.6]
    (Photo,58.51)     [58.5]
    (Avg,63.59)       [63.6]
};
\addplot+[fill=oursColor,
    nodes near coords, point meta=explicit symbolic,
    every node near coord/.append style={color=black!85},
] coordinates {
    (Cora,69.32)      [69.3\\{\color{deltaColor}\tiny+2.1}]
    (WikiCS,76.84)    [76.8\\{\color{deltaColor}\tiny+3.8}]
    (Instagram,57.26) [57.3\\{\color{deltaColor}\tiny+1.7}]
    (Photo,69.89)     [69.9\\{\color{deltaColor}\tiny+11.4}]
    (Avg,68.33)       [68.3\\{\color{deltaColor}\tiny+4.7}]
};

\end{groupplot}

\node[anchor=south, font=\scriptsize, yshift=4pt]
  at (group c1r1.north) {%
    \tikz\draw[fill=baseColor,draw=none](0,0)rectangle(1.6ex,1.6ex);~Base LLM\quad
    \tikz\draw[fill=oursColor,draw=none](0,0)rectangle(1.6ex,1.6ex);~TRN-R1-Zero\quad
    {\color{deltaColor}\textbf{+$\Delta$}\,gain}%
};

\node[anchor=north, font=\scriptsize, yshift=-3pt]
  at (group c1r1.south) {(a) Llama-3.2-3B-Instruct};
\node[anchor=north, font=\scriptsize, yshift=-3pt]
  at (group c1r2.south) {(b) Llama-3.1-8B-Instruct};
\node[anchor=north, font=\scriptsize, yshift=-20pt]
  at (group c1r3.south) {(c) Qwen2.5-14B-Instruct};
\end{tikzpicture}